\RequirePackage{amsmath}
\documentclass[runningheads]{llncs}
\usepackage{amsmath}
\usepackage{graphicx}
\usepackage{siunitx}
\usepackage{cite}
\DeclareMathOperator*{\argmax}{argmax}
\usepackage{boldline,multirow}
\usepackage{color}
\usepackage{soul} 
\usepackage[colorlinks=true,citecolor=blue,urlcolor=black]{hyperref}

\begin{document}
%
\title{Overfitting of neural nets under class imbalance: Analysis and improvements for segmentation}
\titlerunning{Overfitting of neural nets under class imbalance}
%

\author{Zeju Li \and Konstantinos Kamnitsas \and Ben Glocker}

\institute{Biomedical Image Analysis Group, Imperial College London, UK\\
\email{\{zeju.li18,konstantinos.kamnitsas12,b.glocker\}@imperial.ac.uk}}
\authorrunning{Zeju~Li, Konstantinos~Kamnitsas, and Ben~Glocker}

\maketitle              
\begin{abstract}
Overfitting in deep learning has been the focus of a number of recent works, yet its exact impact on the behavior of neural networks is not well understood. This study analyzes overfitting by examining how the distribution of logits alters in relation to how much the model overfits. Specifically, we find that when training with few data samples, the distribution of logit activations when processing unseen test samples of an under-represented class tends to shift towards and even across the decision boundary, while the over-represented class seems unaffected. In image segmentation, foreground samples are often heavily under-represented. We observe that sensitivity of the model drops as a result of overfitting, while precision remains mostly stable. Based on our analysis, we derive asymmetric modifications of existing loss functions and regularizers including a large margin loss, focal loss, adversarial training and mixup, which specifically aim at reducing the shift observed when embedding unseen samples of the under-represented class. We study the case of binary segmentation of brain tumor core and show that our proposed simple modifications lead to significantly improved segmentation performance over the symmetric variants.

\end{abstract}
\section{Introduction}

Convolutional neural networks (CNNs) work exceptionally well when trained with sufficiently large and representative data. When only small amounts of training data is available, overfitting can become a critical issue. CNNs may memorize specific patterns of the training data, leading to poor generalization at test time. Image segmentation is particularly prone to overfitting, as the generation of high-quality expert annotations is tedious and time-consuming. Contributing to the problem is the often severe class imbalance where the foreground class (say tumor) is heavily under-represented in the training samples. Class ratios of 1:10 and lower are typical. To alleviate class imbalance, one may use data augmentation, change the sampling weights per class, add information in the loss function \cite{lin2017focal}, or adopt multi-stage approaches with candidate proposals and background suppression \cite{valindria2018small}. We argue that the key connection between class-imbalance and overfitting of under-represented foreground class has not been investigated sufficiently. Many methods have been proposed for deep models to improve generalization and prevent overfitting including specific loss functions \cite{liu2016large}, data mixing \cite{zhang2017mixup} or learning based augmentation \cite{goodfellow2014explaining}. However, most of these techniques were proposed for general image classification where class imbalance is not specifically addressed. It also remains unclear how these techniques exactly affect the model and to our knowledge this has not been explored in great detail. In this study, we shed new light on the problem of overfitting in the presence of class imbalance aiming to improve segmentation performance.

To explore the effects of overfitting on the model behavior, we investigate how the distribution of activations in the last network layer (\emph{logits}) changes for a model trained with different amounts of training data. We notice that samples of the under-represented class at test time tend to be mapped towards and across the decision boundary, while the mapping of training and test samples of the over-represented class remains stable. This leads to a tendency for the under-represented class losing sensitivity at test time. We argue this is a consequence of class imbalance and overfitting to the few training samples of the under-represented class. Current solutions aiming to make different classes separate better do not address this imbalance and may even reduce the performance in such imbalanced settings, as we show empirically. Based on our analysis, we propose asymmetric modifications of those techniques to steer their effect to tackle the problem of class imbalance, showing promising results for image segmentation with small amounts of training data.

\section{Analysis}

In order to investigate the influence of overfitting, we train convolutional neural networks using different amounts of data with strong class imbalance. We conduct experiments on brain tumor core \cite{bakas2017advancing} and small organ segmentation (data from~\cite{xu2015efficient}). For tumor segmentation, we test on 95 cases and train separate models using 190 (100\%), 95 (50\%), 38 (20\%), 19 (10\%) and 10 cases (5\% of full training set). For organ segmentation, we test on 10 cases and train models using 20 (100\%) and 5 cases (25\% of training set). We use the DeepMedic~\cite{kamnitsas2017efficient} architecture for all our experiments.

\begin{figure}[t]
\centering
\includegraphics[width=\textwidth]{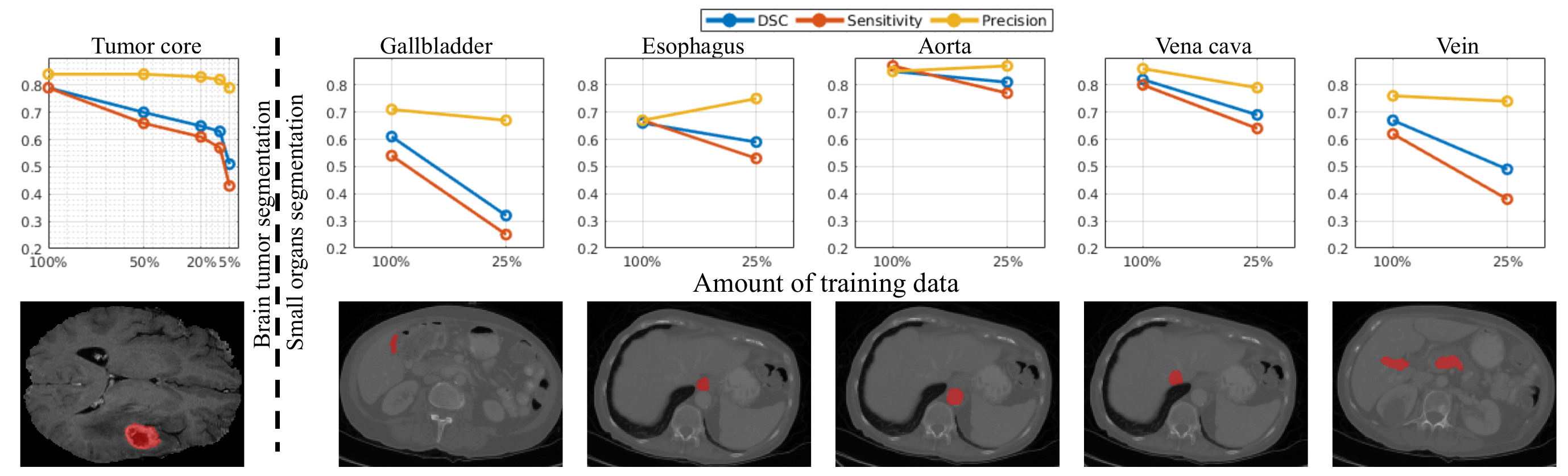}
\caption{Performance on brain tumor core and small organs segmentation with varying amounts of training data. With less training data, performances decline due to the drastic reduction of sensitivity, while precision is retained.}
\label{fig1}
\end{figure}

Results are shown in Fig.~\ref{fig1}. With less training data, we observe a clear decline of segmentation performance on test data but similar or increase of performance on training data, as expressed by the DSC metric (defined as $ DSC=2\frac{\text{sensitivity} \cdot \text{precision}}{\text{sensitivity} + \text{precision}}$). Precision remains largely stable, while we observe that overfitting causes reduced sensitivity. We also observe this behavior in other tasks where foreground classes are under-represented. Our findings reveal that \emph{models that overfit to training data have a bias to under-segment the under-represented class on unseen test data}.

\begin{figure}
\centering
\includegraphics[width=\textwidth]{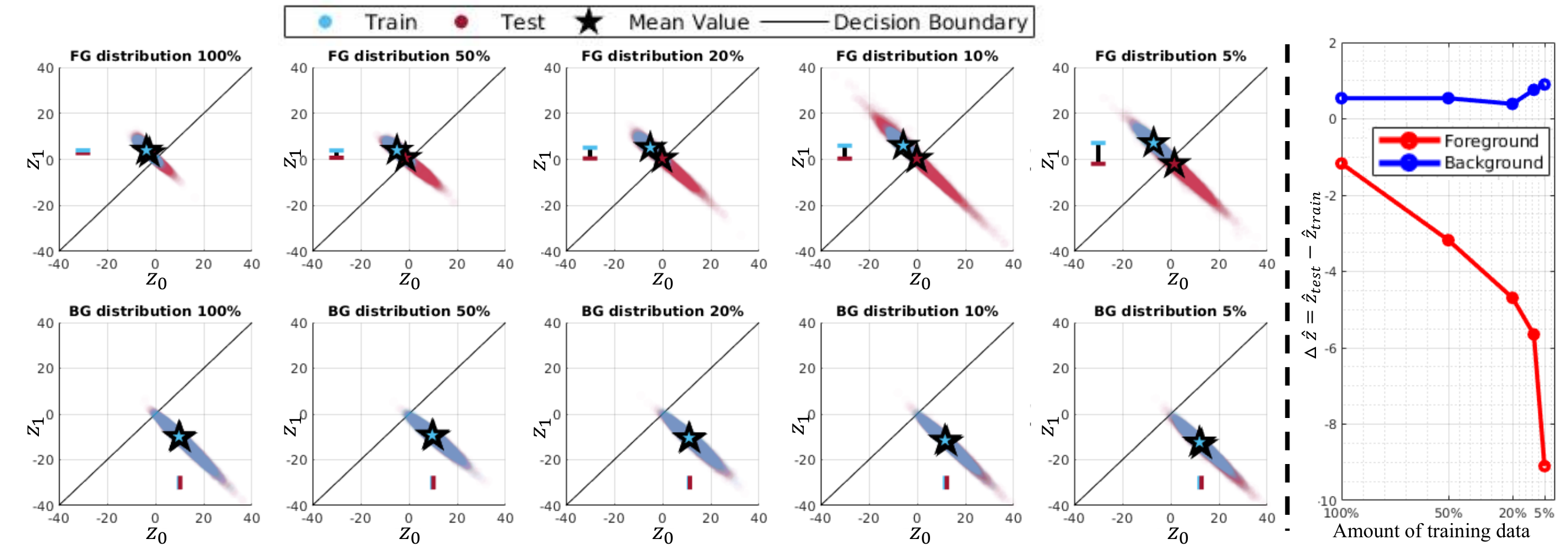}
\caption{(Left part) Activations of the classification layer (logit $z_{0}$ for background, logit $z_{1}$ for foreground) when processing (top) foreground (FG) and (bottom) background (BG) samples, with different amounts of training data. CNN maps training and testing samples of the background class to similar logit values. However, mean activation for testing data shifts significantly for the foreground class towards and sometimes across the decision boundary. (Right part) The shift of mean value of logits observed when processing training and testing data ($ \Delta \hat{z} = |\hat{z}_{test}| - |\hat{z}_{train}| $). }
\label{fig2}
\end{figure}

Delving deeper into analysing this behavior, we monitor the logits when the network processes foreground and background samples of training and unseen test data (cf. plots in Fig.~\ref{fig2}). For simplicity, we focus on the problem of binary segmentation of tumor core for the rest of this paper. We observe that the distributions of logit activations when processing background samples from the training and test sets tend to be similar. However, the distribution of logit activations when processing foreground samples shift significantly towards or even across the decision boundary which causes false negatives. This shift tends to increase for models that overfit more (trained with less training data), leading to our previous observation that sensitivity decreases drastically when models overfit causing the model to under-segment the structures of interest.

We argue that the above behavior is a combined effect of class imbalance and overfitting. The background covers large part of the image and is a relatively heterogeneous class in many tasks. For example, a CNN ``sees'' very different patterns when processing different parts of the brain through its receptive field. Thus to minimize the training cost for even little data, the network has to learn relatively generic filters. Subsequently these filters will also map unseen data appropriately and no shift between the embeddings of training and test samples is observed. In contrast, the appearance of small foreground structures may be easier to memorize within a network, as there are only limited ways the CNN can view them through its receptive field. Even if the structure is complex, a set of case-specific filters could enable memorization. These filters tailored for each training case are suboptimal for new unseen data, yielding poor generalization. As their evaluation on new data does not match well the underlying patterns it leads to activations of smaller magnitude, causing the observed distribution shift\footnote{The dot product between a filter and a signal is highest when these match perfectly.}. As a result of class imbalance and overfitting, the CNN tends to underperform for under-represented classes. The shift of the foreground logit distribution is the cause of the drastic decrease of sensitivity and under-segmentation in case of overfitting. However, previous loss functions and regularization techniques that aim to prevent overfitting do not take this behavior into account and are unable to improve segmentation in this setting. Here, we introduce new asymmetric variants that aim at reducing the shift of the under-represented classes leading to significant improvements.

\section{Method}

\begin{figure}[t]
\centering
\includegraphics[width=\textwidth]{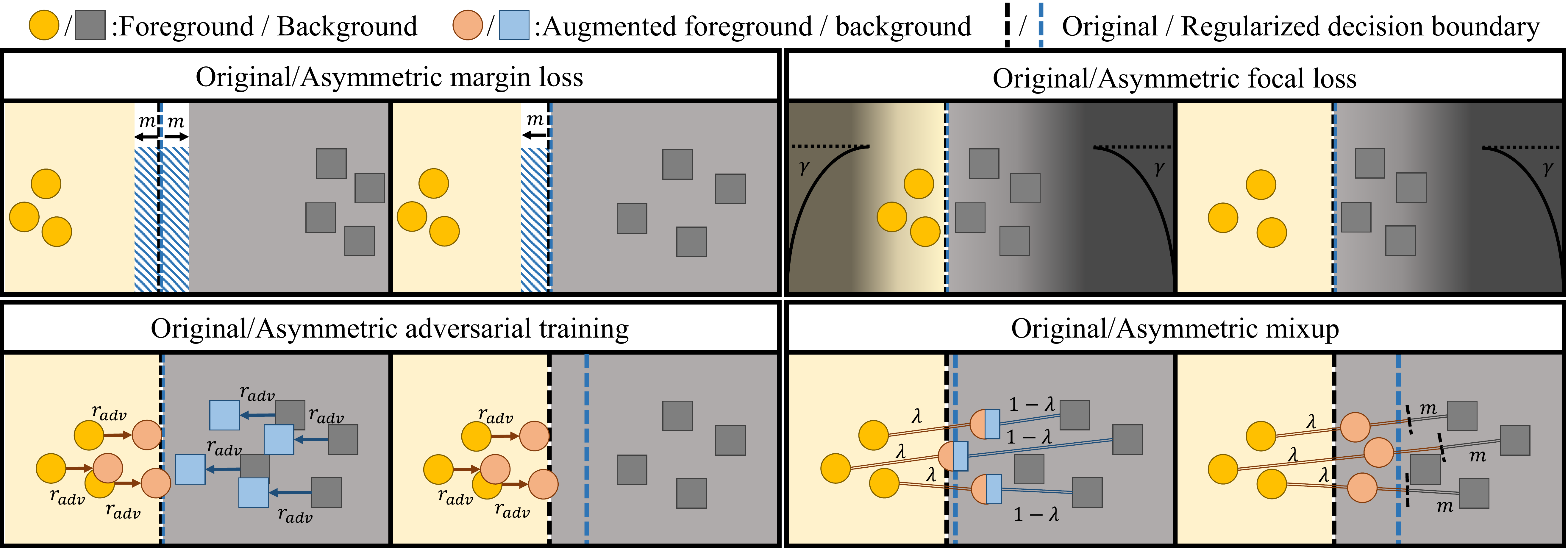}
\caption{The illustration of the proposed asymmetric modifications for the existing four techniques. We make the logit activations of foreground class far away from the decision boundary by setting bias for the foreground class in different ways.} \label{fig3}
\end{figure}

Based on our observations about the behavior of CNNs, we modify previous loss functions and training strategies to prevent distribution shift of logit activations. Specifically, we add a bias for the under-represented class to tackle overfitting under class imbalance. Although the original techniques were proposed for different purposes, our modifications have a common goal: \emph{keep the logit activations of the under-represented class away from the decision boundary}. Even if the logit of a foreground sample shifts towards the decision boundary, its prediction is likely to remain correct (cf. Fig.~\ref{fig3}).

\subsection{Asymmetric large margin loss}

We start with a basic CNN segmentation model. Typically, the softmax function and cross entropy are computed on the logit to make predictions, with a loss as:

\begin{equation}
    L_{CE}(x_{i},y_{i}) = -\frac{1}{N}\sum_{c} y_{i:c}\log(\frac{\mathrm{e}^{z_{c}}}{\sum_{k} \mathrm{e}^{z_{k}}}),
    \label{eq:CE}
\end{equation}

where \emph{N} is the number of training samples, \emph{z$_{c}$} is the network's logit output of the \emph{c}th class, \emph{x$_{i}$} is the input image and \emph{y$_{i}$} is its corresponding one-hot label.

The large margin loss was proposed for increasing the Euclidean distances between logits for different classes to learn discriminative features \cite{liu2016large}. Symmetrically, it is implemented by adding bias on the logits of every class:

\begin{equation}
    L_{M}(x_{i},y_{i}) = -\frac{1}{N}\sum_{c} y_{i:c}\log(\frac{\mathrm{e}^{z_{c}-m}}{\mathrm{e}^{z_{c}-m}+\sum_{k\neq c} \mathrm{e}^{z_{k}}}),
    \label{eq:margin}
\end{equation}

where \emph{m} is the margin. Although the large margin loss encourages the model to map different classes away from each other, the decision boundary remains in the center. According to our findings, the bias of class imbalance causes shifts of unseen foreground samples towards the background class. To mitigate this, a regularizer should move the decision boundary closer to the background class. Therefore, our asymmetric modification pushes the foreground class away from the decision boundary by only setting the margin for the rare class \emph{r}:

\begin{equation}
    \hat{L}_{M}(x_{i},y_{i}) = -\frac{1}{N} y_{i:r}\log(\frac{\mathrm{e}^{z_{r}-m}}{\mathrm{e}^{z_{r}-m}+\sum_{k\neq r} \mathrm{e}^{z_{k}}})-\frac{1}{N}\sum_{c\neq r} y_{i:c}\log(\frac{\mathrm{e}^{z_{c}}}{\sum_{k} \mathrm{e}^{z_{k}}}).
    \label{eq:modified margin}
\end{equation}

\subsection{Asymmetric focal loss}
The focal loss was proposed for small object detection by reducing the loss of well-classified samples and focusing on samples which are near the decision boundary \cite{lin2017focal}. It adds attenuation inside the loss function based on the logit activations:

\begin{equation}
    L_{Focal}(x_{i},y_{i}) = -\frac{1}{N}\sum_{c} (1-\frac{\mathrm{e}^{z_{c}}}{\sum_{k} \mathrm{e}^{z_{k}}})^\gamma y_{i:c}\log(\frac{\mathrm{e}^{z_{c}}}{\sum_{k} \mathrm{e}^{z_{k}}}),
    \label{eq:Focal}
\end{equation}

where $\gamma$ is the hyper-parameter to control the focus. Symmetric focal loss prevents logits from being too large and makes every class stay near the decision boundary. However, it also makes it easy for the unseen foreground samples to shift across the decision boundary. Therefore, we remove the loss attenuation for the foreground class to keep it away from the decision boundary:

\begin{equation}
    \hat{L}_{Focal}(x_{i},y_{i}) = -\frac{1}{N} y_{i:r}\log(\frac{\mathrm{e}^{z_{r}}}{\sum_{k} \mathrm{e}^{z_{k}}})-\frac{1}{N}\sum_{c\neq r} (1-\frac{\mathrm{e}^{z_{c}}}{\sum_{k} \mathrm{e}^{z_{k}}})^\gamma y_{i:c}\log(\frac{\mathrm{e}^{z_{c}}}{\sum_{k} \mathrm{e}^{z_{k}}}).
    \label{eq:modified focal}
\end{equation}

\subsection{Asymmetric adversarial training}
Adversarial training was proposed to learn a robust classifier by training with difficult samples which can break the correct predictions in a significant way~\cite{goodfellow2014explaining}:
\begin{equation}
    L_{adv}(x_{i},y_{i}) = L_{CE}(x_{i},y_{i}) + L_{CE}(x_{i}+l \cdot \frac{d_{adv}}{\|d_{adv}\|_2},y_{i}),
    \label{eq:Adversarial training}
\end{equation}

\begin{equation}
    \textrm{with}\quad  d_{adv} = \argmax_{d;\|d\|<\epsilon}L_{CE}(x_{i}+d,y_{i}).
    \label{eq:Adversarial direction}
\end{equation}

Here, \emph{d$_{adv}$} is the direction of generated adversarial samples, \emph{l} and $\epsilon$ are the magnitude and the range of the adversarial perturbations. Similar to the large margin loss, symmetric adversarial training preserves the decision boundary and causes difficulties for unseen foreground samples, which tends to shift towards background class. Our proposed asymmetric adversarial training aims to produce a larger space between foreground class and the decision boundary:

\begin{equation}
    \hat{d}_{adv} = \argmax_{d;\|d\|<\epsilon}L_{CE}(x_{i}+d,y_{i})\Big|_{y_{i}=r}.
    \label{eq:Adversarial direction2}
\end{equation}

\subsection{Asymmetric mixup}
Mixup is a simple yet effective data augmentation algorithm to improve generalization by generating extra training samples by using the linear combinations of pairs of images and their labels \cite{zhang2017mixup}:

\begin{equation}
    L_{mixup}(x_{i},y_{i},x_{j},y_{j}) = L_{CE}(x_{i},y_{i}) + L_{CE}(\tilde{x_{i}},\tilde{y_{i}}),
    \label{eq:mixup}
\end{equation}

where ($\tilde{x_{i}}$, $\tilde{y_{i}}$) are the generated training sample:

\begin{equation}
    \tilde{x_{i}} = \lambda x_{i}+(1-\lambda)x_{j}, \quad \tilde{y_{i}} = \lambda y_{i}+(1-\lambda)y_{j}.
    \label{eq:mix_x}
\end{equation}

Here, $\lambda$ is randomly selected based on a beta distribution, (\emph{x$_{j}$}, \emph{y$_{j}$}) is another random training sample. Mixup regularizes the model by centering the decision boundary between classes which helps little in our setting. Different from the original mixup, which generates samples with soft labels, our modification generates hard labels by regarding augmented samples which are near to foreground samples just as foreground class. By doing this, asymmetric mixup can keep the decision boundary away from the foreground class and increase the area of foreground logit distribution. This prevents unseen under-presented samples from shifting across the decision boundary. Specifically, the mixed image \emph{$\tilde{x_{i}}$} which has a certain distance from background class, is taken as a foreground sample:

\begin{equation}
\hat{\tilde{y_{i}}}=\left\{
\begin{array}{rcl}
y_{i}         &      & {\text{if}\ (\lambda > m \quad \textrm{and} \quad y_{i} = r)  \quad \textrm{or} \quad y_{i} = y_{j} },\\
y_{j}       &      & {\text{if}\ (1-\lambda > m \quad \textrm{and} \quad y_{j} = r) \quad \textrm{or} \quad y_{i} = y_{j} },\\
0       &      & \textrm{otherwise}
\end{array} \right.
\label{eq:mix_yRE}
\end{equation}

where \emph{m} is the margin to guarantee that the augmented samples are not too close to background samples and still belong to positive samples.

\section{Experiments}

We demonstrate the effect of our proposed modifications for the case of brain tumor core binary segmentation. The hyper-parameters are kept the same for the original baselines and our modified techniques. The quantitative segmentation results are summarized in Table~\ref{tab2}. For baseline experiments, we show that simply changing the objective function to DSC (which is a mix of sensitivity and precision) does not improve the performance. Increasing the weight of tumor samples (from 50\% to 80\%) leads to even more overfitting and decreased performance. Our proposed modifications lead to improvements in all experiments. Specifically, the original large margin loss and mixup sometimes decrease performance, while our modifications boost the performance to a large extent. Focal loss and adversarial training can be effective when data is very little, where our modifications seem to further improve the sensitivity. We also demonstrate that our four methods can be integrated into a single model. The combination of the four modified techniques further improves results.

\begin{table}[t]
\centering
\caption{Evaluation of the tumor core segmentation with different amounts of training data and different techniques to counter overfitting. }\label{tab2}
\newsavebox{\tablebox}
\begin{lrbox}{\tablebox}
\begin{tabular}{p{46mm}<{\centering}|c|c|c||c|c|c||c|c|c||c|c|c}
\hlineB{3}
\multirow{2}{*}{Method} & \multicolumn{3}{c||}{5\% training} & \multicolumn{3}{c||}{10\% training} & \multicolumn{3}{c||}{20\% training} & \multicolumn{3}{c}{50\% training}  \\ 
 &  DSC & SENS &  PRC &   DSC & SENS &  PRC &   DSC & SENS &  PRC &   DSC & SENS &  PRC  \\
\hlineB{2}
Vanilla - CE &  0.51  & 0.43 &  0.79 &  0.63  & 0.57 &  0.82 &  0.65  & 0.61 &  0.83 & 0.70 & 0.66 & 0.84 \\
Vanilla - DSC &  0.48  & 0.39 &  0.82 &  0.59  & 0.52 &  0.83 &  0.65  & 0.59 &  0.83 & 0.67 & 0.63 & 0.85 \\
Vanilla - CE - 80\% tumor &  0.46  & 0.38 &  0.77 &  0.62  & 0.55 &  0.79 &  0.66  & 0.61 &  0.82 & 0.69 & 0.65 & 0.83 \\
\hline
Large margin loss &  0.46  & 0.38 &  0.78 &  0.61  & 0.54 &  0.82 &  0.67  & 0.63 &  0.83 & 0.67 & 0.63 & 0.86 \\
Asymmetric large margin loss &  0.55 & 0.51 &  0.76 &  0.64  & 0.58 &  0.84 &  0.68  & 0.64 &  0.82 & 0.69 & 0.66 & 0.85  \\
\hline
Focal loss &  0.54  & 0.46 &  0.78 &  0.63  & 0.56 &  0.82 &  0.65  & 0.61 &  0.83 & 0.67 & 0.63 & 0.85 \\
Asymmetric focal loss &  0.57 & 0.53 &  0.74 &  0.66  & 0.63 &  0.79 &  0.68  & 0.67 &  0.79 & 0.71 & 0.72 & 0.80  \\
\hline
Adversarial training &  0.53  & 0.46 &  0.79 &  0.62  & 0.56 &  0.83  &  0.65  & 0.60 &  0.83 & 0.66 & 0.62 & 0.85\\
Asymmetric adversarial training &  0.57 & 0.52 &  0.75 &  0.64  & 0.59 &  0.80 &  0.68  & 0.64 &  0.83 & 0.71 & 0.69 & 0.82\\
\hline
Mixup &  0.50 &  0.42  & 0.78  &  0.61  & 0.55 &  0.81 &  0.65  & 0.60 &  0.82 & 0.67 & 0.63 & 0.86 \\
Asymmetric mixup &  0.59 &  0.58  & 0.71  &  0.69  & 0.66 &  0.79 &  0.71  & 0.69 &  0.81 & 0.71 & 0.69 & 0.84 \\
\hline

Asymmetric combination &  \textbf{0.62}  & 0.66 &  0.71  &  \textbf{0.71} & 0.73 &  0.75 &  \textbf{0.72}  & 0.74 &  0.79 & \textbf{0.73} & 0.77 & 0.78 \\

\hline
\end{tabular}
\end{lrbox}
\scalebox{0.805}{\usebox{\tablebox}}
\end{table}

\begin{figure}[t]
\centering
\includegraphics[width=\textwidth]{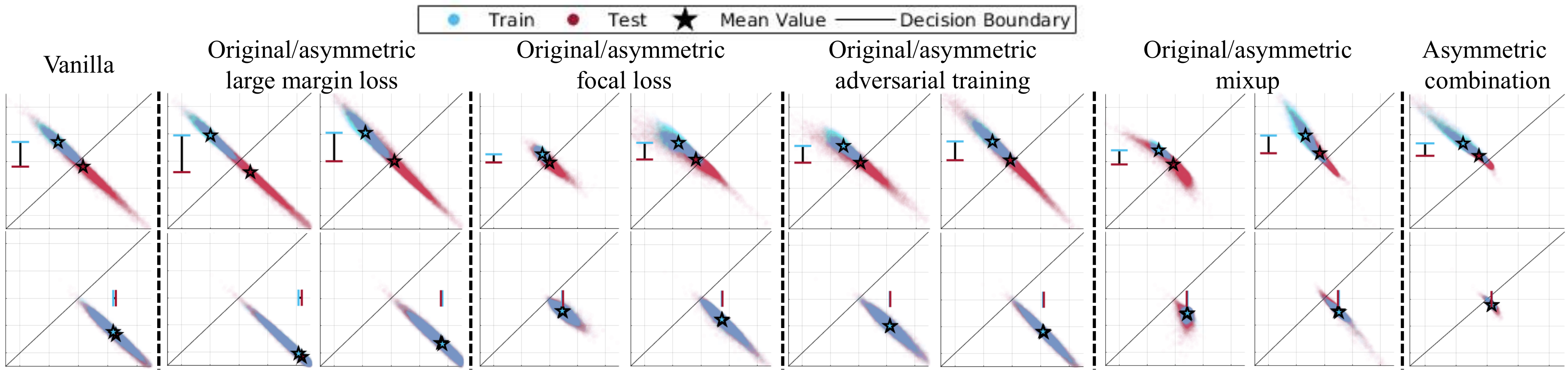}

\caption{Activations of the classification layer when processing (top) foreground and (bottom) background samples, using 5\% training data. Asymmetric modifications lead to better separation of the logits of unseen foreground samples.} \label{fig4}
\end{figure}

The effect of all four techniques on the logit distribution is shown in Fig.~\ref{fig4}. The original large margin loss and adversarial training try to push samples from different classes far from each other, however, the logits of unseen data remain in the center around the decision boundary and thus the predictions are not improved. With our modifications, only the logits of foreground samples are pushed away and the unseen foreground logits tend to remain positive. The original focal loss encourages the network to prevent the logits of each class from staying too far from the decision boundary. However, it allows foreground logits to remain near the decision boundary which can yield false negative predictions. Our asymmetric focal loss removes the constraints of foreground samples. Original mixup encourages the symmetric distributions of different classes but does not consider class imbalance. Asymmetric mixup exploits the embedding space based on the relationship between samples to generate foreground samples and make the decision boundary stay near the background class. This leads overall the biggest improvement by increasing the region for the foreground logit distribution and reduce logit shift of unseen foreground samples.

\section{Conclusion}

In this paper, we analyze overfitting of neural networks under class imbalance. We find that when processing unseen under-represented samples, the logit activations tend to shift towards the decision boundary, thus the sensitivity drops. We derive asymmetric variants for existing loss functions and regularization techniques to prevent overfitting, showing promising results. We expect findings to extend naturally to multi-class problems, which is further investigated in future work. We further believe that our logit distribution plots can be a valuable tool for practitioners to study overfitting and other behavior of different models.

\section*{Acknowledgements}
ZL is grateful for a China Scholarship Council (CSC) Imperial Scholarship. This project has received funding from the European Research Council (ERC) under the European Union's Horizon 2020 research and innovation programme (grant No 757173, project MIRA, ERC-2017-STG) and EPSRC (EP/R511547/1).
%
%
%
%
\bibliographystyle{splncs04}
\bibliography{reference}

\end{document}